\documentclass{article}

\usepackage{epsf}

\usepackage{amssymb}
\usepackage{amsmath}
\usepackage{latexsym}
\usepackage{comment}

\usepackage{graphicx}
\usepackage{comment}
\usepackage{textcomp}
\usepackage{float}
\usepackage{siunitx}
\usepackage{subfigure}
\usepackage{setspace}
\usepackage{comment}

\usepackage[utf8]{inputenc}
\usepackage[frenchb]{babel}

\usepackage{hyperref}
\usepackage{subfiles}

\usepackage{tikz}
\usetikzlibrary{calc}
\usetikzlibrary{positioning}
\tikzset{
    >=stealth,
    hair lines/.style={line width = 0.05pt, lightgray},
    true scale/.style={scale=#1, every node/.style={transform shape}},
}

\DeclareMathOperator\FFNN{FFNN}

\DeclareMathOperator\logsoftmax{log-softmax}
\DeclareMathOperator\argmax{argmax}
\DeclareMathOperator\GRU{GRU}
\DeclareMathOperator\fGRU{\overrightarrow{GRU}}
\DeclareMathOperator\bGRU{\overleftarrow{GRU}}

\begin{document}

\title{Seq2Biseq: Bidirectional Output-wise Recurrent Neural Networks for Sequence Modelling}

\author{Marco Dinarelli$^{(1)}$, Loïc Grobol$^{(2,3)}$\\[18pt]
\small (1) LIG, Bâtiment IMAG - 700 avenue Centrale - Domaine Universitaire de Saint-Martin-d’Hères\\ 
\small (2) Lattice CNRS, 1 rue Maurice Arnoux, 92120 Montrouge, France\\
\small (3) ALMAnaCH Inria, 2 rue Simone Iff, 75589 Paris, France\\
}

\maketitle

\begin{abstract}
During the last couple of years, Recurrent Neural Networks (RNN) have reached state-of-the-art performances on most of the sequence modelling problems. In particular, the \emph{sequence to sequence} model and the neural CRF have proved to be very effective in this domain. In this article, we propose a new RNN architecture for sequence labelling, leveraging gated recurrent layers to take arbitrarily long contexts into account, and using two decoders operating forward and backward. We compare several variants of the proposed solution and their performances to the state-of-the-art. Most of our results are better than the state-of-the-art or very close to it and thanks to the use of recent technologies, our architecture can scale on corpora larger than those used in this work.
\end{abstract}

\section{Introduction}
\label{sec:Intro}

Sequence modelling is an important problem in NLP, as many NLP tasks can be modelled as sequence-to-sequence decoding.
Among them are POS tagging, chunking, named entity recognition \cite{Collobert-2011-NLP-1953048.2078186}, Spoken Language Understanding (SLU) for human-computer interactions \cite{demori08-SPM}, and also machine translation \cite{Sutskever-2014-SSL-2969033.2969173,DBLP-journals-corr-BahdanauCB14}.

In other cases, NLP tasks can be decomposed, at least in principle, in several subtasks, the first of which is a sequence modelling problem.
For instance, syntactic parsing can be performed by applying syntactic analysis to POS-tagged sentences \cite{Collins-1997-TGL}; coreference chain detection \cite{Soon2001,Ng2002,Grouin.etAL:I2B2:2011} can be decomposed into mention detection and coreferent mention linking; and structured named entity detection \cite{Grouin.etAL:I2B2:2011,dinarelli2012-eacl,DINARELLI-ROSSET:OCR-NER:LREC2012}, can be done by first detecting simple entity components then combining them to construct complex tree-shaped entities.

Most of these tasks can also be performed by a single model: either as a joint architecture like the joint model for POS tagging and syntactic analysis from \cite{Rush-2012-IPP-2390948.2391112} or with a fully end-to-end model like the one developed by \cite{D17-1018} for coreference detection.
In any case, these models still include at some point a sequence modelling module that could be improved by studying successful models for the related sequence labelling tasks.

This is even more true for neural models, since designing a single complex neural architecture for a complex problem may indeed lead to sub-optimal learning.
For this reason, it may be more desirable to train a sequence labelling model alone at first and to learn to perform the other steps using the pre-trained parameters of the first step's model, as is done for instance when using pre-trained lexical embeddings in a downstream model \cite{lample2016neural,Ma-Hovy-ACL-2016}.
In that case, care must be taken to avoid too unrelated downstream tasks that could lead to \emph{Catastrophic forgetting} \cite{kemker2018measuring}, though some hierarchical multi-task architectures have proven successful \cite{N18-1172}.

Finally, \cite{DBLP-journals-corr-VinyalsKKPSH14} has shown that it is possible to model syntactic analysis as a sequence labelling problem by adapting a \emph{Seq2seq} model. As a consequence, we could actually design a unified multi-task learning neural architecture for a large class of NLP problems, by recasting them as sequence decoding tasks.

Recurrent Neural Networks (RNNs) hold state-of-the-art results in many NLP tasks, and in particular in sequence modelling problems \cite{lample2016neural,Ma-Hovy-ACL-2016,dinarelli_hal-01553830,D17-1018}.
Gated RNNs such as GRU and LSTM are particularly effective for sequence labelling thanks to an architecture that allows them to use long-range information in their internal representations \cite{werbos-bptt,Hochreiter-1997-LSTM,Cho-2014-GatedRecurrentUnits}.

In this paper we focus our work to searching for more effective neural models for sequence labelling tasks such as POS tagging or Spoken Language Understanding (SLU).
Several very effective solutions already exist for these problems, in particular the sequence-to-sequence model \cite{Sutskever-2014-SSL-2969033.2969173} (\emph{Seq2seq} henceforth), the \emph{Transformer} model \cite{46201}, and the whole family of models using a neural CRF layer on top of one or several LSTM or GRU layers \cite{Hochreiter-1997-LSTM,Cho-2014-GatedRecurrentUnits,lample2016neural,Ma-Hovy-ACL-2016,Vukotic.etal_2016,LSTM-CNN-NER-2015,huang2015bidirectional}.

We propose an alternative neural architecture to those mentioned above.
This architecture uses GRU recurrent layers as internal memory capable of taking into account arbitrarily long contexts of both input (words and characters), and output (labels).
Our architecture is a variant of the \emph{Seq2seq} model where two different decoders are used instead of only one of the original architecture. The first decoder goes backward through the sequence, outputting label predictions, using the hidden states of the encoder and its own previous hidden states and label predictions as input.
The second decoder is a more standard forward decoder that uses the hidden states of the encoder, the hidden states and \emph{future} predictions generated by the backward decoder and its own previous hidden states and predictions to output labels.
We name this architecture \emph{Seq2biseq}, as it generates output sequences from output-wise bidirectional, global decisions.

Our work is inspired by previous work published in \cite{dinarelli_hal-01553830,Dupont-etAl-LDRNN-CICling2017,2016:arXiv:DinarelliTellier:NewRNN,DinarelliTellier:RNN:CICling2016}, where bidirectional output-wise decisions were taken using a simple recurrent network.
A similar idea, called \textit{deliberation network}, has been proposed in \cite{NIPS2017_6775}, where however two forward decoders were used. In this respect we believe that using a backward decoder for the first pass may encode more different, expressive information for the second, forward pass.
Our architecture takes global decisions like a \emph{LSTM+CRF} model \cite{lample2016neural} thanks to the use of the two decoders. These take global context into account on both sides of a given position of the input sequence.

We compare our solution with state-of-the-art models for SLU and POS-tagging in particular the models described in \cite{dinarelli_hal-01553830,Dupont-etAl-LDRNN-CICling2017} and in \cite{lample2016neural}.
In order to make a direct comparison, we evaluate our models on the same tasks: a French SLU task provided with the MEDIA corpus \cite{Bonneau-Maynard2006-media}, and the well-known task of POS-tagging of the Wall Street Journal portion of the Penn Treebank \cite{Marcus93buildinga}.

Our results are all reasonably close to the state of the art, and most of them are actually better.

The paper is organized as follows: in the next section we describe the state-of-the-art of neural models for sequence labelling.
In the section~\ref{sec:GRU-IRNN} we describe the neural model we propose in this paper, while in the section~\ref{sec:eval} we describe the experiments we performed to evaluate our models.
We draw our conclusions in the section~\ref{sec:Conclusions}

\section{State of the Art}
\label{sec:SOTA}

The two main neural architectures used for sequence modelling are the \emph{Seq2seq} model \cite{Sutskever-2014-SSL-2969033.2969173} and a group of models where a neural CRF output layer is stacked on top of one or several LSTM or GRU layers \cite{Hochreiter-1997-LSTM,Cho-2014-GatedRecurrentUnits,lample2016neural,Ma-Hovy-ACL-2016,Vukotic.etal_2016,LSTM-CNN-NER-2015,huang2015bidirectional}.

The \emph{Seq2seq} model, also known as \emph{encoder-decoder}, uses a first module to encode the input sequence as a single vector $c$.
In the version of this model proposed in \cite{Sutskever-2014-SSL-2969033.2969173} $c$ is the hidden state of the encoder after seeing the whole input sequence.
A second module decodes the output sequence using its previous predictions and $c$ as input.

The subsequent work of \cite{DBLP-journals-corr-BahdanauCB14} extends this model with an attention mechanism.
This mechanism provides the decoder with a dynamic representation of the input that depends on the decoding step, which proved to be more efficient for translating long sentences.

This mechanism has also been turned out to be effective for other NLP tasks \cite{D17-1018,DBLP-journals-corr-KimDHR17,simonnet-hal-01433202}.

Concerning models using a neural CRF output layer \cite{Ma-Hovy-ACL-2016,lample2016neural}, a first version was already described in \cite{Collobert-2011-NLP-1953048.2078186}.
These solutions use one or more recurrent hidden layers to encode input items (words) in context. Earlier simple recurrent layers like \emph{Elman} and \emph{Jordan} \cite{Elman90findingstructure,jordan-serial}, which showed limitations for learning long-range dependencies \cite{Bengio-1994-RNN-Learning-Difficulty}, have been replaced by more sophisticated layers like LSTM and GRU \cite{Hochreiter-1997-LSTM,Cho-2014-GatedRecurrentUnits}, which reduced such limitations by using gates.

In this type of neural models, a first representation of the prediction is computed with a local output layer.
In order to compute global predictions with a CRF neural layer, the \emph{Viterbi} algorithm is applied over the sequence of local predictions \cite{Collobert-2011-NLP-1953048.2078186,Mesnil-RNN-2015}.

A more recent neural architecture for sequence modelling is the \emph{Transformer} model \cite{46201}. This model use an innovative deep non-recurrent neural architecture, relying heavily on attention mechanisms \cite{DBLP-journals-corr-BahdanauCB14} and skip connections \cite{Bengio03aneural} to overcome limitations of recurrent networks in propagating the learning signal over long paths. The Transformer model has been designed for computational efficiency reasons, but it captures long-range contexts with multiple attention mechanisms (multi-head attention) applied to the whole input sequence. Skip-connections guarantee that the learning signal is back-propagated effectively to all the network layers.

Concerning previous works on the same tasks used in this work, namely MEDIA \cite{Bonneau-Maynard2006-media} and the Penn Treebank (WSJ) \cite{Marcus93buildinga}, several publications have been produced starting from $2007$ (MEDIA) and $2002$ (WSJ) \cite{raymond07-luna,dinarelli09:Interspeech,Hahn.etAL-SLUJournal-2010,Dinarelli2010.PhDThesis,dinarelli2011:emnlp,Dinarelli.etAl-SLU-RR-2011}, applying several different models like \emph{SVM} and \emph{CRF} \cite{Vapnik98-book,lafferty01-crf}.
Starting from 2013 several works also focused on neural models. At first simple recurrent networks have been used \cite{RNNforSLU-Interspeech-2013,RNNforLU-Interspeech-2013,Vukotic.etal_2015}. In the last few years also more sophisticated models have been studied \cite{SLUwithLSTM-NN-IEEEWshop-2014,Vukotic.etal_2016,dinarelli_hal-01553830}.

\section{The Seq2biseq Neural Architecture}
\label{sec:GRU-IRNN}

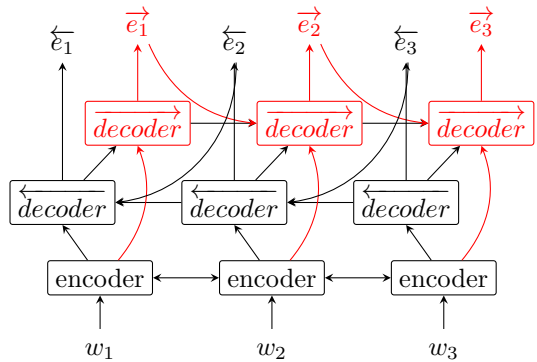
\begin{figure}
\center
\begin{tikzpicture}
	\def\words{$w_1$, $w_2$, $w_3$}
	\def\foutputs{$\overrightarrow{e_1}$, $\overrightarrow{e_2}$, $\overrightarrow{e_3}$}
	\def\boutputs{$\overleftarrow{e_1}$, $\overleftarrow{e_2}$, $\overleftarrow{e_3}$}
	
	\begin{scope}[local bounding box=net]
	\foreach \w [count=\wi from 1, remember=\wi as \lastwi] in \words {
	\ifnum\wi>1
            	\node[right=6.5em of w\lastwi.center, text height=1.5ex, text depth=0.25ex, anchor=center] (w\wi) {\w};
            \else
            	\node[text height=1.5ex, text depth=0.25ex] (w\wi) {\w};
            \fi
	}
	\foreach \w [count=\wi from 1, remember=\wi as \lastwi] in \words {
		\draw[->] (w\wi) -- +(0, 2em) node[draw, anchor=south, rounded corners=1pt, inner sep=0.3em] (ew\wi) {encoder};
		\ifnum\wi>1
			\draw[<->] (ew\lastwi) -- (ew\wi);
		\fi
	}
	
	\foreach \w [count=\wi from 1, remember=\wi as \lastwi] in \words {
		\draw[->] (ew\wi) -- +(-0.5, 2em) node[draw, anchor=south, rounded corners=1pt, inner sep=0.3em] (bdec\wi) {$\overleftarrow{decoder}$};
		\ifnum\wi>1
			\draw[->] (bdec\wi) -- (bdec\lastwi);
		\fi
	}
	
	\foreach \w [count=\wi from 1, remember=\wi as \lastwi] in \words {
		\draw[->,red] (ew\wi) to[bend right=30] +(+0.5, 5em) node[draw, anchor=south, rounded corners=1pt, inner sep=0.3em] (fdec\wi) {$\overrightarrow{decoder}$};
		\draw[->] (bdec\wi) -- (fdec\wi);
		\ifnum\wi>1
			\draw[->] (fdec\lastwi) -- (fdec\wi);
		\fi
	}
	
	\foreach \y [count=\yi from 1, remember=\yi as \lastyi] in \foutputs {
            	\draw[red,->] (fdec\yi) -- +(0, 3em) node[text height=1.5ex, text depth=0.25ex, anchor=south] (fy\yi) {\y};
            	\ifnum\yi>1
            		\draw[red,->] (fy\lastyi) to[bend right=30] (fdec\yi);
            	\fi
	}
	
	\foreach \y [count=\yi from 1, remember=\yi as \lastyi] in \boutputs {
		\draw[->] (bdec\yi) -- +(0, 5.3em) node[text height=1.5ex, text depth=0.25ex, anchor=south] (by\yi) {\y};
	}
	
	\draw[->] (by3) to[bend left=42] (bdec2.east);
	\draw[->] (by2) to[bend left=42] (bdec1.east);

        \end{scope}

\end{tikzpicture}
    \caption{Overall network structure}\label{fig:network-structure}
\end{figure}

As an alternative to the \emph{Seq2seq} and \emph{LSTM+CRF} neural models for sequence labelling, we propose in this paper a new neural architecture inspired from the original \emph{Seq2seq} model and from models described in \cite{dinarelli_hal-01553830,Dupont-etAl-LDRNN-CICling2017}.
Figure~\ref{fig:network-structure} shows the overall architecture.

Our architecture is similar to the \emph{Seq2seq} model in that we use modules to encode a long-range context on the output side similar to the decoder of the \emph{Seq2seq} architecture.
The similarity with respect to models described in \cite{dinarelli_hal-01553830,Dupont-etAl-LDRNN-CICling2017} is the use of a bidirectional context on the output side in order to take into account previous, but also future predictions for the current model decision. Future predictions are computed by an independent decoder which processes the input sequence backward.

Our architecture extends the \emph{Seq2seq} original model through the use of an additional backward decoder that allows taking into account both past and future information at decoding time.
Our architecture also improves the models described in \cite{dinarelli_hal-01553830,Dupont-etAl-LDRNN-CICling2017} since it uses more sophisticated layers to model long-range contexts on the output side, while previous models used fixed-size windows and simple linear hidden layers.
Thanks to these modifications our model makes predictions informed by a global distributional context, which approximates a global decision function.
We also improve the character-level word representations by using a similar solution to the one proposed in \cite{Ma-Hovy-ACL-2016}.

Our neural architecture is based on the use of GRU recurrent layers at word, character and label levels.
GRU is an evolution of the LSTM recurrent layer which has often shown better capacities to model contextual information \cite{Cho-2014-GatedRecurrentUnits,Vukotic.etal_2016}.

In order to make notation clear, in the following sections, bidirectional GRU hidden layers are noted $\GRU$, while we use $\fGRU$ and $\bGRU$ for a forward and backward hidden layer respectively.
For the output of these layers we use respectively $h_{w_i}$, $\overrightarrow{h_{e_i}}$ and $\overleftarrow{h_{e_i}}$, with a letter as index to specialize the GRU layer for a specific input (e.g. $w$ for the GRU layer used for words, $e$ for labels, or entities, and so on), and an index $i$ to indicate the index position in the current sequence.
For example $\overleftarrow{h_{e_i}}$ is the backward hidden state, at current position ($i$), of the GRU layer for labels.
The models described in this work always use as input words, characters and labels. Their respective embedding matrices are all noted $E_x$, with $x$ denoting the different input unit types (e.g. $E_w$ is the embedding matrix for words), and their dimensions $D_x$.

\subsection{Character-level Representations}
\label{subsec:char_rep}

The character-level representation of words was computed at first as in \cite{Ma-Hovy-ACL-2016}, substituting a GRU to the LSTM layer:
the characters $c_{w, 1}, \dots, c_{w, n}$ of a word $w$ are first represented as a sequence $S_c(w)$ of $n$ $D_c$-dimensional embeddings. These are fed to the $\GRU_c$ layer. The final state $h_c(w)$ is kept as the character level representation of $w$.

We improved this module so that it generates a character-level representation using all the hidden states generated by $\GRU_c$:
\begin{equation}\label{eq:charrep}
    \begin{aligned}
        S_c(w) &= (E_{c}(c_{w, 1}), \dots, E_c(c_{w, n})) \\
        (h_c(c_{w, 1}), \dots, h_c(c_{w, n})) &= \GRU_c(S_{c}(w), h_0^c) \\
        h_c(w) &= \FFNN( Sum( h_c(c_{w, 1}), \dots, h_c(c_{w, n}) ) )
    \end{aligned}
\end{equation}
$\FFNN$ is again a general, possibly multi-layer Feed-Forward Neural Network with non-linear activation functions. This new architecture was inspired by \cite{46201}, where $\FFNN$s were used to extract deeper features at each layer.

Preliminary experiments have shown that this character-level representation is more effective than the one inspired by the work of \cite{Ma-Hovy-ACL-2016}.

\subsection{Word-level Representations}
\label{subsec:word_rep}

Words are first mapped into embeddings, then the embedding sequence is processed by a $\GRU_{w}$ bidirectional layer.
Using the same notation as for characters, a sequence of words $S = w_1, \dots, w_N$ is converted into embeddings $E_w(w_i)$ with $1\leq i \leq N$.
We denote $S_i = w_1, \dots, w_i$ the sub-sequence of $S$ up to the words $w_i$.
In order to augment the word representations with their character-level representations, and to use a single distributed representation, we concatenate the character-level representations $h_{c}(w_i)$ (eq.~\ref{eq:charrep}) to the word embeddings before feeding the $\GRU_w$ layer with the whole sequence. Formally:
\begin{equation}\label{eq:lex-rep}
    \begin{aligned}
        S_w &= (E_{w}(w_{1}), \dots, E_{w}(w_{N})) \\
        S^{lex} &= ([E_{w}(w_{1}), h_{c}(w_1)], \dots, [E_{w}(w_{N}), h_{c}(w_N)]) \\
        h_{w_i} &= \GRU_w(S_i^{lex}, h_{w_{i-1}})
    \end{aligned}
\end{equation}
Where we used $S_w$ for the whole sequence of word embeddings generated from the word sequence $S$.

In the same way, $S^{\mathrm{lex}}$ is the sequence obtained concatenating word embeddings and character-level representations, which constitute the lexical-level information given as input to the model.
$[~]$ is the matrix (or vector) concatenation, and we also used the notation $S_i^{\mathrm{lex}}$ for the sub-sequence of $S^{\mathrm{lex}}$ up to position $i$.

\subsection{Label-level Representations}
\label{subsec:label_rep}

In order to obtain label representations encoding long-range contexts, we use a $\GRU$ hidden layer also on label embeddings.
We apply first a backward step on label embeddings in order to compute representations that will be used as future label predictions, or right context, in the following forward step.
Using the same notation as used previously, we have:
\begin{equation}\label{eq:label-bw}
    \overleftarrow{h_{e_i}} = \bGRU_e(E_l(e_{i+1}), \overleftarrow{h_{e_{i+1}}})
\end{equation}
for $i = N \dots 1$.
We note that here we use the label on the right of the current position, $e_{i+1}$, $e_{i}$ is not known at time step $i$.

The hidden state $\overleftarrow{h_{e_{i+1}}}$ is the hidden state computed at previous position in the backward step, thus associated to the label on the right of the current label to be predicted. In other words we interpret $\overleftarrow{h_{e_{i}}}$ as the right context of the (unknown) label $e_i$, instead of as the in-context representation of $e_i$ itself, and similarly for $\overleftarrow{h_{e_{i+1}}}$.
The right context of $e_i$, $\overleftarrow{h_{e_i}}$, is used to predict $e_i$ at time step $i$.

In the same way, we compute the representation of the left context of the label $e_i$ by scanning the input sequence forward, which gives:
\begin{equation}\label{eq:label-fw}
    \overrightarrow{h_{e_i}} = \fGRU_{e}(E_l(e_{i-1}), \overrightarrow{h_{e_{i-1}}})
\end{equation}
for $i = 1 \dots N$.
The neural components described so far are already sufficient to build rich architectures.
However, we believe that the information from the lexical context is useful not only to disambiguate the current word in-context, but also to disambiguate the contextual representations used for label prediction.
Indeed, in sequence labelling labels only provide abstract lexical or semantic information. It thus seems reasonable to think that they are not sufficient to effectively encode features in the label context representations $\overleftarrow{h_{e_i}}$  and $\overrightarrow{h_{e_i}}$.

For this reason, we add to the input of the layers $\bGRU_{e}$ and $\fGRU_{e}$ the lexical hidden representation $h_{w_i}$ computed by the $\GRU_{w}$ layer.
Taking this into account, the computation of the right context for the current label prediction becomes:
\begin{equation}\label{eq:label-bw-le}
    \overleftarrow{h_{e_i}} = \bGRU_{e}([h_{w_i}, E_l(e_{i+1})], \overleftarrow{h_{e_{i+1}}})
\end{equation}
The computation of the left context is done in a similar way.

This modification makes the layers $\bGRU_{e}$ and $\fGRU_{e}$ in our architecture similar to the decoder of a \emph{Seq2seq} architecture \cite{Sutskever-2014-SSL-2969033.2969173}.
The modules $\bGRU_{e}$ and $\fGRU_{e}$ are indeed like two decoders from an architectural point of view, but also they encode the contextual information in the same way using gated recurrent layers.

However, the full architecture differs from a traditional \emph{Seq2seq} model by the use of an additional decoder, capable of modelling the right label context, while the original model used a single decoder, modelling only the left context.
The idea of using two decoders is inspired mainly by the evidence that both left and right output-side contexts are equally informative for the current prediction.

Another difference with respect to the \emph{Seq2seq} model is that the $\bGRU_{e}$ and $\fGRU_{e}$ layers have access to the lexical-level hidden states $h_{w_i}$.
This allows these layers to take the current lexical context into account and is thus more adapted to sequence labelling than using the same representation of the input sentence for all the positions, which is the solution of the original \emph{Seq2seq} model.

As we mentioned above, the \emph{Seq2seq} model has been improved with an attention mechanism \cite{DBLP-journals-corr-BahdanauCB14}, which is another way to provide the model with a lexical representation focusing dynamically on different parts of the input sequence depending on the position $i$.
This attention mechanism has also proved to be efficient for sequence labelling, and it might be that our architecture could benefit from it too, but this is out of our scope for this article and we leave it for future work.\footnote{This is currently in progress}

We can motivate the use of the lexical information $h_{w_i}$ in the decoders $\bGRU_{e}$ and $\fGRU_{e}$ with complex systems theory considerations, as suggested in \cite{2017-NoRNN}.
\cite{Holland-1999-ECO-520475} state that a complex system, either biological or artificial, is not equal to the sum of its components.
More precisely, the behaviour of a complex system evolves during its existence and shows the emergence of new functionalities, which can not be explained by simply considering the system's components individually.
\cite{RePEc-wop-safiwp-93-11-070} qualitatively characterizes the evolution of a complex system's behaviour with three different types of adaptation, two of which are particularly interesting in the context of this work and can be concisely named \emph{aggregation} and \emph{specialization}.

In the first, several components of the system adapt in order to become a single \emph{aggregated} component from a functioning point of view.
In \emph{specialization}, several initially identical components of the system adapt to perform different functionalities.
These adaptations may take place at different unit levels, a neuron, a simple layer, or a whole module.

The most evident cases of \emph{specialization} are the gates of the LSTM or GRU layers \cite{Cho-2014-GatedRecurrentUnits}, as well as the attention mechanism \cite{DBLP-journals-corr-BahdanauCB14}.
Indeed, the $\mathbf{z}$ and $\mathbf{r}$ gates of a GRU recurrent layer are defined in the exact same way, with the same number of parameters, and they use exactly the same input information.

However, during the evolution of the system — that is, during the learning phase — the $\mathbf{r}$ gate adapts (specialises) to become the reset gate, which allows the network to forget the past information, when it is not relevant for the current prediction step.
In the same way, the $\mathbf{z}$ gate becomes the equivalent of the input gate of a LSTM, which controls the amount of input information that will affect the current prediction.

In our neural architecture we can observe \emph{aggregation}: the layers $\bGRU_{e}$ and $\fGRU_{e}$ adapt at the whole layer level, they become like gates which filter label-level information that is not useful for the current prediction.
In the same way as the input to gates of GRU or LSTM is made of current input and previous hidden state, the input to the $\bGRU_{e}$ and $\fGRU_{e}$ layers is made of lexical level and previous label level information, both needed to discriminate the abstract semantic information provided by the labels alone.
We will show in the evaluation section the effectiveness provided by this choice.

While both of the two decoders used in our models are equivalent to the decoder of the original \emph{Seq2seq} architecture, we believe it is interesting to analyse the contribution of each piece of information given as input to this component, which we will show in the evaluation section.

\subsection{Output Layer}
\label{subsec:output}

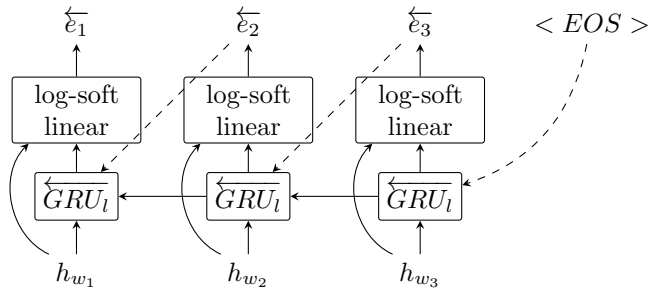
\begin{figure}
\center
\begin{tikzpicture}
	\def\words{$h_{w_1}$, $h_{w_2}$, $h_{w_3}$}
	\def\outputs{$\overleftarrow{e_{1}}$, $\overleftarrow{e_{2}}$, $\overleftarrow{e_{3}}$}
	
	\begin{scope}[local bounding box=net]
	\foreach \w [count=\wi from 1, remember=\wi as \lastwi] in \words {
	\ifnum\wi>1
            	\node[right=6.5em of w\lastwi.center, text height=1.5ex, text depth=0.25ex, anchor=center] (w\wi) {\w};
            \else
            	\node[text height=1.5ex, text depth=0.25ex] (w\wi) {\w};
            \fi
	}
	
	\foreach \w [count=\wi from 1, remember=\wi as \lastwi] in \words {
		\draw[->] (w\wi) -- +(0, 2em) node[draw, anchor=south, rounded corners=1pt, inner sep=0.3em] (gruw\wi) {$\overleftarrow{GRU_l}$};
		\ifnum\wi>1
			\draw[<-] (gruw\lastwi) -- (gruw\wi);
		\fi
	}
	
	\foreach \w [count=\wi from 1, remember=\wi as \lastwi] in \outputs {
		\draw[->] (gruw\wi) -- +(0, 2em) node[draw, text width=1.5cm, text centered, anchor=south, rounded corners=1pt, inner sep=0.3em] (lin\wi) {log-soft linear};
		\draw[->] (w\wi) to[bend left=55] (lin\wi);
	}
	
	\foreach \w [count=\wi from 1, remember=\wi as \lastwi] in \outputs {
		\draw[->] (lin\wi) -- +(0, 2.5em) node[text height=1.5ex, text depth=0.25ex, anchor=south] (o\wi) {\w};
		\ifnum\wi>1
			\draw[->,dashed] (o\wi) -- (gruw\lastwi);
		\fi
	}
	
	\node[right=6.5em of o3.center, text height=1.5ex, text depth=0.25ex, anchor=center] (eos) {$<EOS>$};
	\draw[->,dashed] (eos) to[bend left] (gruw3);

        \end{scope}

\end{tikzpicture}
\caption{Structure of the backward decoder}\label{pic:back-dec}
\end{figure}

Once all pieces of information needed to predict the current label are computed, the output of the backward step is computed as follows:
\begin{equation}\label{eq:backward-model}
    \begin{aligned}
        o_{bw} &= W_{bw} [h_{w_{i}}, \overleftarrow{h_{e_i}}] + b_{bw} \\
        e_{i} &= \argmax(\logsoftmax(o_{bw}))
    \end{aligned}
\end{equation}
We start the backward step using a conventional symbol (\texttt{<EOS>}) as end-of-sentence marker.
We repeat the backward step prediction for the whole input sequence. The process is shown in figure~\ref{pic:back-dec}.

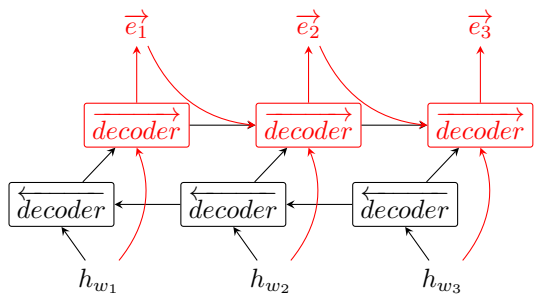
\begin{figure}
\center
\begin{tikzpicture}
	\def\words{$h_{w_1}$, $h_{w_2}$, $h_{w_3}$}
	\def\outputs{$\overrightarrow{e_1}$, $\overrightarrow{e_2}$, $\overrightarrow{e_3}$}
	
	\begin{scope}[local bounding box=net]
	\foreach \w [count=\wi from 1, remember=\wi as \lastwi] in \words {
	\ifnum\wi>1
            	\node[right=6.5em of w\lastwi.center, text height=1.5ex, text depth=0.25ex, anchor=center] (ew\wi) {\w};
            \else
            	\node[text height=1.5ex, text depth=0.25ex] (ew\wi) {\w};
            \fi
	}
	
	\foreach \w [count=\wi from 1, remember=\wi as \lastwi] in \words {
		\draw[->] (ew\wi) -- +(-0.5, 2em) node[draw, anchor=south, rounded corners=1pt, inner sep=0.3em] (bdec\wi) {$\overleftarrow{decoder}$};
		\ifnum\wi>1
			\draw[->] (bdec\wi) -- (bdec\lastwi);
		\fi
	}
	
	\foreach \w [count=\wi from 1, remember=\wi as \lastwi] in \words {
		\draw[->,red] (ew\wi) to[bend right=30] +(+0.5, 5em) node[draw, anchor=south, rounded corners=1pt, inner sep=0.3em] (fdec\wi) {$\overrightarrow{decoder}$};
		\draw[->] (bdec\wi) -- (fdec\wi);
		\ifnum\wi>1
			\draw[->] (fdec\lastwi) -- (fdec\wi);
		\fi
	}
	
	\foreach \y [count=\yi from 1, remember=\yi as \lastyi] in \outputs {
            	\draw[red,->] (fdec\yi) -- +(0, 3em) node[text height=1.5ex, text depth=0.25ex, anchor=south] (fy\yi) {\y};
            	\ifnum\yi>1
            		\draw[red,->] (fy\lastyi) to[bend right=30] (fdec\yi);
            	\fi
	}

        \end{scope}

\end{tikzpicture}
\caption{High-level schema of the forward pass}\label{pic:for-dec}
\end{figure}

This allows to have all the pieces of information needed to predict the current label in the forward step, at character and word level, but also at right and left label context level, with respect to the current position to be labeled:
\begin{equation}\label{eq:bidirectional-model}
    \begin{aligned}
        o_i &= W_o [\mathbf{\overrightarrow{h_{e_i}}}, h_{w_i}, \mathbf{\overleftarrow{h_{e_i}}}] + b_o \\
        e_i &= \argmax(\logsoftmax(o_i))
    \end{aligned}
\end{equation}
A high-level schema of the forward pass is shown in figure~\ref{pic:for-dec}.

The $\logsoftmax$ function computes log-probabilities and it is thus suited for the loss-function used to learn the model described in the next section.

We note that the forward decoder is in fact a bidirectional decoder, as it uses both backward and forward hidden states $\overrightarrow{h_{e_i}}$ and $\overleftarrow{h_{e_i}}$ for the current prediction.

The hypothesis motivating the architecture of our neural models is the following: gated hidden layers such as LSTM and GRU can keep relatively long contexts in memory and to extract from them the information that is relevant to the current model prediction.
This is supported by the findings in recent works, such as \cite{P18-2116}, which shows that most of the modelling power of gated RNN comes from their ability to compute at each step a context-dependent weighted sum on their inputs, in a way that is akin to dynamical attention mechanism.
As an immediate consequence, we think that using such hidden layers is an effective way to keep in memory a relatively long context on the output item level, that is labels, as well as on the input item level, that is words, characters and possibly other information.

An alternative, non-recurrent architecture, the Transformer model \cite{46201} has been proposed with the goal of using attention mechanisms to overcome the learning issues of RNN in contexts where the learning signal has to back-propagate through very long paths.
However, the recent work of \cite{Dehghani2018UniversalT} shows that integrating a concept of recurrence in Transformers can improve their performances in some contexts.
This leads us to believe that recurrence is a fundamental feature for neural architectures for NLP and all of the domains where data are sequential by nature.

As a side note, the main features of the Transformer model - the multi-head attention mechanism and the skip connections \cite{46201} - could in principle be integrated into our architecture.
Investigations of the costs and benefits of such additions is left for future work.

Finally, while the decision function of our model remains local, its decisions are informed by global information at the word, character and label level thanks to the use of long-range contexts encoded by the GRU layers.
In that sense, it can be interpreted as an approximation of a global decision function and provides a viable alternative to the use of a CRF output layer \cite{lample2016neural,Ma-Hovy-ACL-2016}.

\subsection{Learning}
\label{subsec:learning}

Our models are learned by minimizing the negative log-likelihood $\mathcal{LL}$ with respect to the data. Formally:
\begin{equation}\label{eq:LL}
    -\mathcal{LL}(\Theta | D) = -\sum_{d=1}^{|D|} \sum_{i=1}^{N_d} \frac{1}{2}(\text{log-p}(\overrightarrow{e_i})+\text{log-p}(\overleftarrow{e_i}))  + \frac{\lambda}{2} \left | \Theta \right |^2 )
\end{equation}
$\text{log-p}(\overrightarrow{e_i})$ and $\text{log-p}(\overleftarrow{e_i})$ are the log-probabilities over predictions of the forward and backward decoders, respectively, we thus strengthen the global character of our model's predictions.
The first sum scans the learning data $D$ of size $|D|$, while the second sum scans each learning sequence $S_d$, of size $N_d$.

Given the relatively small size of the data we use for the evaluation, and the relatively high complexity of the models proposed in this paper, we add a $L_2$ regularization term to the cost function with a $\lambda$ coefficient.
The cost-function is minimized with the \emph{Back-propagation Through Time} algorithm (BPTT) \cite{werbos-bptt}, provided natively by the \emph{Pytorch} library (see section~\ref{subsec:settings}).

\section{Evaluation}
\label{sec:eval}

\subsection{Data}
\label{subsec:data}

We evaluate our models on two tasks, one of Spoken Language Understanding (SLU), and one of POS tagging, namely \emph{MEDIA} and \emph{WSJ} respectively.
These tasks have been widely used in the literature \cite{Vukotic.etal_2015,Vukotic.etal_2016,dinarelli_hal-01553830,Ma-Hovy-ACL-2016,2018-ijcai-lstm-ldrnn} and allow thus for a direct comparison of results.

\textbf{The French MEDIA corpus} \cite{Bonneau-Maynard2006-media} was created for the evaluation of spoken dialog systems in the domain of hotel information and reservation in France.
It is made of $1~250$ human-machine dialogs acquired with a \textit{Wizard-of-OZ} approach, where \num{250} users followed \num{5} different reservation scenarios.

Data have been manually transcribed and annotated with domain concepts, following a rich ontology.
Semantic components can be combined to build relatively complex semantic classes.\footnote{For example, the label \texttt{localisation} can be combined with the components \texttt{ville} (city), \texttt{distance-relative} (relative-distance), \texttt{localisation-relative-générale} (general-relative-localisation), \texttt{rue} (street), etc.}

Statistics on the training, development and test data of the MEDIA corpus are shown in table~\ref{tab:MEDIAStats}.
The MEDIA task can be modelled as a sequence labelling task by segmenting concepts over words with the BIO formalism \cite{Ramshaw95-BIO}.
An example of sentence with its semantic annotation is shown in table~\ref{tab:ATIS-MEDIA-exemple}.
For exhaustive, we also show some word-classes available for this task, allowing models for a better generalization.
However, our model does not use these classes, as explained in section~\ref{subsec:settings}.

\textbf{The English corpus Penn Treebank} \cite{Marcus93buildinga}, and in particular the section of the corpus corresponding to the articles of Wall Street Journal (WSJ), is one of the most known and used corpus for the evaluation of models for sequence labelling.

The task consists of annotating each word with its Part-of-Speech (POS) tag. We use the most common split of this corpus, where sections from \num{0} to \num{18} are used for training ($38~219$ sentences, $912~344$ tokens), sections from \num{19} to \num{21} are used for validation ($5~527$ sentences, $131~768$ tokens), and sections from \num{22} to \num{24} are used for testing ($5~462$ sentences, $129~654$ tokens).

\begin{table}[t]
	    \centering
	    \scriptsize
	    \begin{tabular}{|ccc|}
	    \hline
	    \multicolumn{3}{|c|}{MEDIA corpus example} \\
	    \textbf{Words} & \textbf{Classes} & \textbf{Labels} \\
	    \hline
        Oui & - & Answer-B \\
        l' & - & BDObject-B \\
        hotel & - & BDObject-I \\
        le & - & Object-B \\
        prix & - & Object-I \\
        à & - & Comp.-payment-B \\
        moins & relative & Comp.-payment-I \\
        cinquante & tens & Paym.-amount-B \\
        cinq & units & Paym.-amount-I \\
        euros & currency & Paym.-currency-B \\
        \hline
	    \end{tabular}
	    \caption{An example of sentence with its semantic annotation and word classes, taken from the French corpus MEDIA. The translation of the sentence in English is ``Yes, the hotel with a price less than fifty euros per night''}
	    \label{tab:ATIS-MEDIA-exemple}
    \end{table}

\begin{table}[t]
\begin{minipage}{1.0\linewidth}
    \centering
    \scriptsize
    \begin{tabular}{|l|rr|rr|rr|}
      \hline
      & \multicolumn{2}{|c|}{Training} & \multicolumn{2}{|c|}{Validation} & \multicolumn{2}{|c|}{Test}\\
      \hline
      \# sentences     &\multicolumn{2}{|c|}{12~908} &\multicolumn{2}{|c|}{1~259}&\multicolumn{2}{|c|}{3~005} \\
      \hline
      \hline
      & \multicolumn{1}{|c}{Words} & \multicolumn{1}{c|}{Concepts} &  \multicolumn{1}{|c}{Words} & \multicolumn{1}{c|}{Concepts} &
      \multicolumn{1}{|c}{Words} & \multicolumn{1}{c|}{Concepts} \\
      \hline
      \# words          & 94~466 & 43~078 & 10~849 & 4~705 & 25~606 & 11~383 \\
      \# dict.         &  2~210 &     99 &    838 &    66 &  1~276 &     78 \\
      \# OOV\%   & --     & --     &  1,33  & 0,02  &  1,39  &  0,04  \\
      \hline
    \end{tabular}
    \caption{Statistics on the French MEDIA corpus}
  \label{tab:MEDIAStats}
  \end{minipage}
\end{table}

\subsection{Experimental settings}
\label{subsec:settings}

In order to keep our architecture as general as possible, we limit our model inputs to the strict word (and character) information available in the raw text data and ignore the additional features available in the MEDIA dataset.

For convenience, the hyperparameters of our system have been tuned by simple independent linear searches on the validation data — rather than a grid search on the full hyperparameters space.

All of the parameters of neural layers are initialised with the Pytorch 0.4.1 default initializers\footnote{Uniform random initialization for the GRU layers and \cite{LeakyReLU-PReLU-2015} initialization for the linear layers.} and trained by SGB with a \num{0.9} momentum for \num{40} epochs on MEDIA, and ADAM optimizer for \num{52} epochs on WSJ, keeping the model that gave the best accuracy on the development data set.

For training, we start with a learning rate of \num{0.125} that we decay linearly after each epoch to end up at \num{0} at the end of the chosen number of training epochs.
Following \cite{PracticalRecommendations-Bengio-2012}, we also apply a random dropout to the embeddings and the output of the hidden layers that we optimized to a rate of \num{0.5}, and $L_2$ regularization to all the parameters with an optimal coefficient of \num{e-4}.

Finally, we have conducted experiments to find the optimal layer sizes, which gave us \num{200}, \num{150} and \num{30} for word, labels and character embeddings respectively, \num{100} for the $\GRU_c$ layer and \num{300} for all the other GRU layers.
Those values are for the MEDIA task; for WSJ only the word embeddings and hidden layer sizes (respectively \num{300} and \num{150}) are different.

In order to reduce the training time, we use mini-batches of size\footnote{Using larger batches is faster but degrades the overall accuracy.} \num{100}.
In the current neural network frameworks, all the sequences in a mini-batch must have the same length, which we enforced at first by padding all of the sentences with the conventional symbol \texttt{<s>} to the length of the longest one.
However this caused two problems: first, there are a few unusually long sentences in the datasets we used, for instance, there is a single sentence of \num{198} words in MEDIA.
Secondly, in order to compute automatically the gradients of the parameters, Pytorch keeps in memory the whole graph of operations performed on the input of the model \cite{paszke2017automatic}, which was far too large for the hardware we used, since for our model, we have to keep track of all the operations at all of the timesteps.

We found two solutions to these problems.
The first was to train on fixed-length, overlapping sub-sequences, or segments\footnote{Shifting each segment one token ahead with respect to the previous}, truncated from the whole sentences, which did not appear to impair the performances significantly and allowed us to avoid more involved solutions such as back-propagation through time with memorization \cite{NIPS2016_6221}.
The second was to cluster sentences by their length. This makes small clusters for unusually long sentences, which fit thus in memory, and big clusters of average-length sentences, which are further split into sub-clusters to have an optimal balance between the learning signals of different clusters, and alleviate us to find adaptive learning rates for different clusters.

In the optimization phase, we found out that the first solution works far better for the MEDIA task.
We believe that this is due to the noisy nature of the corpus (speech transcription), and to its relatively small size
Using fixed-length segments reduces the amount of noise the network must filter, while the fact that segments shift and overlap makes the network more robust, as it can see any token as the beginning of a segment, which in turns helps overcoming scarcity of the dataset.
This robustness is not needed when using bigger amount of grammatically well-formed textual data, like the WSJ corpus.
Indeed the two solutions gave similar results on this corpus, we thus preferred sentence clusters which is a more intuitive solution and may better fit bigger data sets.

After performing these optimization on the development set for each task, we kept the best models and evaluated them on the corresponding test sets, which we report and discuss in the next section.

All of our development and experiments were done on \emph{2,1 GHz Intel Xeon E5-2620} CPUs and \emph{GeForce GTX 1080} GPUs.\footnote{1600 MHz, 2560 cores}.

\subsection{Results}
\label{subsec:results}

Results presented in this section on the MEDIA corpus are means over \num{10} runs, while results on the WSJ corpus are obtained in a single run, as it seems the most common practice.\footnote{We can note that results over different runs on the WSJ have a very small variation, less or equal to \num{0.01} accuracy points}

Concerning the MEDIA task, since the model selection during the training phase is done based on the accuracy on the development data, we show accuracy in addition to F1 measure and Concept Error Rate (CER) as it is common practice in the literature on this task.
F1 measure is computed with the script made available to the community for the \emph{CoNLL} evaluation campaign.\footnote{\url{https://github.com/robertostling/efselab/blob/master/3rdparty/conlleval.perl}}. CER is computed by Levenshtein alignment between reference annotation and model hypothesis, with an algorithm much similar to the one implemented in the \emph{sclite} toolkit.\footnote{\url{http://www1.icsi.berkeley.edu/Speech/docs/sctk-1.2/sclite.htm}}

Since our model is similar to \emph{Seq2seq} model, but it uses two decoders, in the remainder of this paper our model will be named \emph{Seq2Biseq}.
The model training is performed using gold labels in the training data, while in test phase the model uses predicted labels to build left and right label-level contexts. This corresponds to the best strategy, according to \cite{RNNforSLU-Interspeech-2013}.

We compare our results to those obtained by running the software developed for \cite{dinarelli_hal-01553830}\footnote{Available upon request at \url{http://www.marcodinarelli.it/software.php}} and tuning its hyperparameters\footnote{The optimal settings being more or less those provided in the original article}.

\begin{table}[t]
\centering
    \begin{tabular}{|l|r|c|c|}

        \hline
        \textbf{Model}  & \textbf{Accuracy} & \textbf{F1 measure} & \textbf{CER} \\ \hline
        \hline
        \multicolumn{4}{|c|}{\textbf{MEDIA DEV}}\\ \hline
        \hline
        Seq2Biseq           				&	89.11		&	85.59	&	11.46		\\
        Seq2Biseq$_{le}$      			&	89.42	&	86.09	&	10.58	\\
        Seq2Biseq$_{le}$ seg-len 15		&	\textbf{89.97}	&	\textbf{86.57}	&	\textbf{10.42}	\\
        \emph{fw}-Seq2Biseq$_{le}$ seg-len 15	&	89.51	&	85.94	&	11.40 \\
        \hline

    \end{tabular}
\caption{Comparison of results on the development data of the MEDIA corpus, with and without the lexical information (Seq2Biseq$_{le}$) as input to the modules $\bGRU_{e}$ and $\bGRU_{e}$}
\label{tab:lex-label-gru}
\end{table}

Concerning our hypothesis about the capability of our models to encode a long-range context, and to filter out useless information with respect to the current labelling decision, we show results of two (sets of) experiments to validate such hypothesis.

In the first one, we compare the results obtained by models with and without the use of the lexical information as input to the decoders $\bGRU_{e}$ and $\bGRU_{e}$ (section~\ref{subsec:label_rep}).
These results are shown in the first two lines of the table~\ref{tab:lex-label-gru}.
The model using the lexical information is indicated with Seq2Biseq$_{le}$ in the table (for \textbf{l}abels and l\textbf{e}xical information).
As we can see in the table, this model obtains much better results than the one not using the lexical information as input to the label decoders.
This confirms that this information helps discriminating the semantic information provided by labels at a given processing step of the input sequence.

In the second experiment, we test the capability of our models to filter out useless semantic information, that is on the label side, for the current labelling decision.
In order to do this, we increase the size of the segments in the learning phase: \num{15} instead of \num{10} by default.
It is important to note that in the context of a SLU task, where input sequences are transcriptions of human speech, using longer segments is possibly risky, since a longer context may be much more noisy even if it is slightly more informative.

Moreover, the models in the literature applied to the MEDIA task and using a fixed-size window to capture contextual information, never use a window wider than \num{3} tokens around the current token to be labelled.
This confirms the difficulty to extract useful information from a longer context.
Results of this experiment are shown in the third line of table~\ref{tab:lex-label-gru}.
Our hypothesis seems to be also valid in this case, as models using segments of length \num{15} obtain better results than those using the default size of \num{10} and this with respect to all the evaluation metrics.

We note that, while the effectiveness of the decoder's architecture of the \emph{Seq2seq} model does not need any more to be proved, these results still provide possibly interesting analyses in the particular context of sequence labelling.\footnote{The \emph{Seq2seq} model has been designed and mainly used for machine translation}

In order to show the advantage provided by the use of two decoders instead of only one like in the original \emph{Seq2seq} model, we show results obtained using only one decoder for the left label-side context in table~\ref{tab:lex-label-gru}
These results are indicated in the table with \emph{fw-Seq2Biseq$_{le}$ seg-len 15} (this model corresponds basically to the original \emph{Seq2seq}). This model is exactly equivalent to our best model \emph{Seq2Biseq$_{le}$ seg-len 15}, the only difference is that it uses only the left label context. As we can see, this model is much less effective than the version using two decoders, which also confirms that the right context on the output side (labels) is very informative.

Our hypothesis concerning the \emph{aggregation} specialization of our model during the learning phase seems also confirmed (section~\ref{subsec:label_rep}).
The fact that the Seq2Biseq$_{le}$ model obtains better results than the simpler model Seq2Biseq tends to confirm the hypothesis.

Indeed, if the model Seq2Biseq$_{le}$ gave more importance to the lexical information than the semantic information given by labels at the input of the decoders $\bGRU_{e}$ and $\fGRU_{e}$, its better results would not have a clear explanation, as both Seq2Biseq$_{le}$ and Seq2Biseq models (table~\ref{tab:lex-label-gru}) use the lexical information separately (indicated with $h_{w_i}$ in the equation~\ref{eq:lex-rep}).

Since the information provided by labels alone is already taken into account by the model Seq2Biseq, we can deduct that the Seq2Biseq$_{le}$ model can extract more effective semantic representations, and this even when we provide it with longer contexts (with segments of size $15$).

\begin{table}[t]
\centering
    \begin{tabular}{|l|r|c|c|c|}

        \hline
        \textbf{Model}  & \textbf{Accuracy} & \textbf{F1 measure} & \textbf{CER} & \textbf{p-value}\\ \hline
        \hline
        \multicolumn{5}{|c|}{\textbf{MEDIA DEV}}\\ \hline
        \hline
        LD-RNN$_{\mathrm{deep}}$				&	89.26 (0.16)	&	85.79 (0.24)	&	10.72 (0.14) & -- \\ \hline
        Seq2Biseq$_{le}$ seg-len 15     	& 89.97 (0.20)	& 86.57 (0.22)	& 10.42 (0.26) & 0.043 \\
        Seq2Biseq$_{\text{2-opt}}$     	& \textbf{90.22} (0.14)	& \textbf{86.88} (0.16)	& \textbf{9.97} (0.24) & -- \\ \hline
        \hline
        \multicolumn{5}{|c|}{\textbf{MEDIA TEST}}\\ \hline
        \hline
        LD-RNN$_{\mathrm{deep}}$				&	89.51 (0.21)	&	87.31 (0.19)	& 10.02 (0.17) & -- \\ \hline
        Seq2Biseq$_{le}$ seg-len 15     	& 89.57 (0.12)	& 87.50 (0.17)	& 10.26 (0.19) & 0.047 \\
        Seq2Biseq$_{\text{2-opt}}$     	& \textbf{89.79} (0.22)	& \textbf{87.69} (0.20)	& \textbf{9.93} (0.28) & -- \\
        \hline

    \end{tabular}
    \caption{Comparison of results obtained on the MEDIA corpus by the system LD-RNN$_{\mathrm{deep}}$, ran by ourselves for this work, and our model Seq2Biseq$_{le}$, using segments of size $15$ (see section~\ref{subsec:settings}).}
\label{tab:ldrnn-vs-gru-ldrnn-clen15}
\end{table}

In another set of experiments, we compared our model with the one proposed in \cite{dinarelli_hal-01553830}, from which we inspired our neural architecture.
We downloaded the software associated to the paper\footnote{Described at \url{http://www.marcodinarelli.it/software.php} and available upon request}, and we ran experiments on the MEDIA corpus in the same conditions as our experiments. We used the deep variant of the model described in \cite{dinarelli_hal-01553830}, LD-RNN$_{\mathrm{deep}}$, which gives the best results on MEDIA. The results of these experiments are shown in the table~\ref{tab:ldrnn-vs-gru-ldrnn-clen15}.
As we can see in the table, on the development data of the MEDIA task (MEDIA DEV), our model is more effective than the LD-RNN$_{\mathrm{deep}}$ of \cite{dinarelli_hal-01553830}, which holds the state-of-the-art on this task.
These gains are also present for the test data (MEDIA TEST), even if they are smaller, and the LD-RNN$_{\mathrm{deep}}$ model is still the more effective in terms of Concept Error Rate (CER).

We would like to underline that we did not perform an exhaustive optimization of all the hyper-parameters.\footnote{This because it takes a lot of time, but more importantly because we believe a good model should give good results without too much effort, otherwise a previous model which already proved comparably effective should be preferred}
As we can see in table~\ref{tab:ldrnn-vs-gru-ldrnn-clen15}, results obtained with the model LD-RNN$_{\mathrm{deep}}$ on the test data are always better than those obtained on the development data. In contrast, our model obtains a worse accuracy, which leads the model selection in the training phase, on the test data. This lack of generalization may indicate a sub-optimal parameter choice or an over-training problem.

In the table~\ref{tab:ldrnn-vs-gru-ldrnn-clen15} we also report standard deviations on the \num{10} experiments (between parentheses), and the results of the significance tests performed on the output of our model and of the model LD-RNN$_{\mathrm{deep}}$. We used the significance test described in \cite{Yeh00moreaccurate}, which applies on the output of the two compared systems, and it is suited for the evaluation metrics used most often in NLP.\footnote{In contrast to several other significance tests, this test doesn't make any assumption on the classes independence, nor on the representative coverage of the sample} We re-implemented the significance test script based on the one described in \cite{sigf06}.\footnote{\url{https://nlpado.de/~sebastian/software/sigf.shtml}}
Our model is compared to the LD-RNN$_{\mathrm{deep}}$ model in terms of F1 measure, which is more constraining than the accuracy and as constraining as the CER.
The result of the significance test is given in the column \emph{p-value} of the table, and it represents the probability that the gain is not significant. Most often the gains are considered significant with a p-value equal or smaller than $0.05$.

\begin{table}[t]
\centering
    \begin{tabular}{|l|rcc|}

        \hline
        \textbf{Model}  & \textbf{Accuracy} & \textbf{F1 measure} & \textbf{CER} \\ \hline
        \hline
        \multicolumn{4}{|c|}{\textbf{MEDIA TEST}}\\ \hline
        \hline
        BiGRU+CRF \cite{dinarelli_hal-01553830}		&	-- 	&	86.69		&	10.13	\\\hline
        LD-RNN$_{\mathrm{deep}}$ \cite{dinarelli_hal-01553830}	& -- 		&	87.36		&	\textbf{9.8}		\\
        LD-RNN$_{\mathrm{deep}}$				&	89.51		&	87.31		&	10.02		\\\hline
        Seq2Biseq$_{le}$ seg-len 15     	&	89.57	&	87.50	&	10.26	\\
        Seq2Biseq$_{\text{2-opt}}$     	& \textbf{89.79}	& \textbf{87.69}	& 9.93 \\ \hline

    \end{tabular}
\caption{Comparison of results on MEDIA with our best models and the best models in the literature}
\label{tab:gru-ldrnn-clen15-vs-SOTA-media}
\end{table}

We ran another set of experiments on the MEDIA task with our best model in order to compare to the best models in the literature on this task, which are those described in \cite{dinarelli_hal-01553830}.
In particular we compared our results to the models using a neural CRF output layer for modelling label sequences and take global decisions.

The results of these experiments are shown in the table~\ref{tab:gru-ldrnn-clen15-vs-SOTA-media}.
In this table we indicate simply with LD-RNN$_{\mathrm{deep}}$ the results obtained in our experiments using the software \emph{LD-RNN}\footnote{\url{http://www.marcodinarelli.it/software.php}}, while we add the reference \cite{dinarelli_hal-01553830} after LD-RNN$_{\mathrm{deep}}$ to indicate that results have been taken directly from the reference.
As we can see, the only new outcome in this table with respect to those already shown in previous tables, is the best CER of $9.8$ obtained by the model LD-RNN$_{\mathrm{deep}}$ published in \cite{dinarelli_hal-01553830}.
These results are obtained however using also the word-classes available with the MEDIA corpus. Our model is still more effective than the others in terms of accuracy and F1 measure, providing thus the new state-of-the-art results on this task.

The experiments performed on the MEDIA task with different variants of our model allowed us to find the best neural architecture for sequence modelling. In order to have a more general view on the effectiveness of our model on the problem of sequence labelling, we performed some experiments of POS tagging on the WSJ corpus, which is a well-known benchmark for sequence labelling, used since more than $15$ years.
In order to show the effectiveness of the model alone, without the impact of any external resources, we performed experiments without using pre-trained embeddings. This is however a quite common practice and can lead to remarkable improvements \cite{Ma-Hovy-ACL-2016}.

On this task we compare to the model \emph{LD-RNN$_{\mathrm{deep}}$} of \cite{dinarelli_hal-01553830}, and to the model \emph{LSTM-CRF} of \cite{Ma-Hovy-ACL-2016}. To the best of our knowledge the latter is one of the rare work on neural models where results are given also without pre-trained embeddings, allowing a direct comparison. The \emph{LSTM-CRF} model is moreover one of the best models on the WSJ corpus when using embeddings pre-trained with GloVe \cite{pennington2014glove}.

The results of the POS tagging task on the WSJ corpus are shown in the table~\ref{tab:comp-WSJ}. As we can see our model obtains the best results among those not using any pre-trained embeddings.
Our results are however worse than those obtained with pre-trained embeddings, which constitute the state-of-the-art on this task.
In this respect, we would like to underline that the overall best results are obtained with a neural model described in \cite{2018-ijcai-lstm-ldrnn}. This model is only slightly better than the \emph{LSTM-CRF} model, which we outperform when not using pre-trained embeddings. Moreover the model proposed in \cite{2018-ijcai-lstm-ldrnn} (\emph{LSTM+LD-RNN} in the table) is very similar to our model.

\begin{table}[t]
\centering
    \begin{tabular}{|l|c|c|}

        \hline
        \textbf{Model}  & \multicolumn{2}{|c|}{\textbf{Accuracy}} \\ \hline
        \hline
        & WSJ DEV & WSJ TEST \\ \hline
        \hline
        LD-RNN$_{\mathrm{deep}}$							& 96.90	& 96.91 \\
        LSTM+CRF \cite{Ma-Hovy-ACL-2016}				& -- 		& 97.13 \\
        Seq2Biseq								& 97.13	& 97.20 \\
        Seq2Biseq$_{\text{2-opt}}$						& \textbf{97.33}	& \textbf{97.35} \\
        \hline
        \hline
        LSTM+CRF + Glove \cite{Ma-Hovy-ACL-2016}		& 97.46 	& 97.55 \\
        LSTM+LD-RNN + Glove \cite{2018-ijcai-lstm-ldrnn}	& -- 		& 97.59 \\ \hline

    \end{tabular}
\caption{Comparison of our model with the model \emph{LD-RNN$_{\mathrm{deep}}$}, and the best models of the literature, on the POS tagging task of the WSJ corpus}
\label{tab:comp-WSJ}
\end{table}

In order to compare our model to the model LD-RNN$_{\mathrm{deep}}$ also in terms of complexity and computation efficiency, we show in the table~\ref{tab:comp-params-train-time} the number of parameters as well as the training time on the MEDIA and WSJ corpora.
For the sake of completeness, we also report the number of parameters of the other models mentioned in this paper. Except for the model \emph{GRU+CRF} for which we took the number of parameters from the reference \cite{dinarelli_hal-01553830} (hidden layers of size $200$), all the other numbers are computed based on the same layer sizes.

We can see in the table~\ref{tab:comp-params-train-time} that the training time for our model is longer than for the model LD-RNN$_{\mathrm{deep}}$ on the MEDIA task.
This is because our neural architecture is quite more complex, and since the corpus is relatively small, we can not fuly take advantage of GPU parallelism.

This is confirmed on the WSJ corpus, where the training time of our model is much smaller than the time needed by the LD-RNN$_{\mathrm{deep}}$ model, despite this corpus is quite bigger than MEDIA.\footnote{The model LD-RNN$_{\mathrm{deep}}$ is coded in Octave, and while it can run on GPUs, this framework is not fully optimized to scale on GPUs}
The time needed for testing are not reported in the table, we can note that they are negligible for both models, as it never exceeded a few minutes

\begin{table}[t]
\centering
    \begin{tabular}{|l|c|c|c|}

        \hline
        \textbf{Model}  & \textbf{\# of parameters} & \multicolumn{2}{|c|}{\textbf{Training time}} \\ \hline
        \hline
        & MEDIA & MEDIA & WSJ \\ \hline
        \hline
        Seq2Biseq$_{le}$					& 2,139,950 & 3h30' & 16h-17h \\
        LD-RNN$_{\mathrm{deep}}$					& 2,551,700 & 1h30' & $>$ 6 days \\
        \hline
        \hline
        GRU+CRF \cite{dinarelli_hal-01553830}	& 2,328,360 & -- & -- \\
        \emph{Seq2seq}						& 1,703,450 & -- & -- \\
        \emph{Seq2seq+Att.}					& 2,244,050 &-- & -- \\ \hline

    \end{tabular}
\caption{Comparison of the neural models proposed or mentioned in this paper, in terms of number of parameters, and of training time for our model and the the model LD-RNN$_{\mathrm{deep}}$}
\label{tab:comp-params-train-time}
\end{table}

While the results described in this paper can be considered satisfactory, considering the complexity of our neural network with respect to the LD-RNN$_{\mathrm{deep}}$ model, we were  surprised to find out that the gains were not larger on the MEDIA task.
At first we thought that our network suffered from overfitting on such a small task, and given the complexity of our network, nothing could be done to solve this problem beyond reducing the total number of parameters.
However, after a quick analysis of the output of our model on the MEDIA development data, we found clear signs revealing that our model was actually ignoring the learning signal coming from the backward decoder (eq.~\ref{eq:backward-model}).

Since our neural network was explicitly designed to take both left and right label-side contexts into account, we thought that the problem was coming from the learning phase. In particular we thought that our model was underfitting due to the problem of \textit{very-long back-propagation paths} described in \cite{46201}, and which motivated the design of the Transformer model, without recurrent layers and with skip connections to enforce the back-propagation of the learning signal.
We adopted a different approach: we applied two different optimizers to the two decoders, one for a negative log-likelihood computed with the output of the backward decoder (only $\text{log-p}(\overleftarrow{e_i})$, see eq.~\ref{eq:backward-model}), and another one for the global negative log-likelihood computed from the output of both forward and backward decoders (see equation~\ref{eq:LL}).
We note that the forward decoder also uses predictions and hidden states of the backward decoder, the second optimizer thus also refines the parameters of the backward decoder with left, forward information.

We ran new experiments in exactly the same conditions as described before, the only difference being that we used these two optimizers.
The final results are reported in table~\ref{tab:ldrnn-vs-gru-ldrnn-clen15} for MEDIA and in the table~\ref{tab:comp-WSJ} for the WSJ, where the model learned using two optimizers is indicated with Seq2Biseq$_{\text{2-opt}}$.

As we can see in the tables, the results improved on both tasks, on both development and test data, and in terms of all the evaluation metrics.
To the best of our knowledge, the results obtained on MEDIA are the best on this task, except for the CER where the model LD-RNN$_{\mathrm{deep}}$ using class features is still the best (9.8 vs. our 9.93 on the test set). Also, the results obtained on the WSJ corpus are the best obtained without any external resource and without pre-trained embeddings. We leave the integration of pre-trained embeddings as future work.

\section{Conclusions}
\label{sec:Conclusions}

In this article, we propose a new neural architecture for sequence modelling heavily based on GRU recurrent hidden layers.
We use these layers to encode long-range contextual information at several levels: words, characters and labels.

Our main contributions are the use of two different decoders for label prediction, one modelling a backward (future, or right) label context, and one for a forward label context.
The combination of the two contexts allow our model to take labelling decisions informed by a global context, approximating a global decision function.
Another contribution is the use of two different optimizers to optimize separately the two decoders.
This improves even further the results obtained on the two evaluation tasks studied in this work.

The results obtained are state-of-the-art on the MEDIA task.
On the POS tagging task of the WSJ corpus, our results are state-of-the-art if we do not consider the models that use pre-trained word embeddings, and still close to the state-of-the-art if we do so.

\section{Acknowledgements}

This work is part of the “Investissements d’Avenir” overseen by the French National Research Agency ANR-10-LABX-0083 (Labex EFL), and is also supported by the ANR DEMOCRAT (Describing and Modelling Reference Chains: Tools for Corpus Annotation and Automatic Processing) project ANR-15-CE38-0008.

\bibliographystyle{splncs2016}
\bibliography{bibliocicling2019}

\begin{thebibliography}{10}

\bibitem{Collobert-2011-NLP-1953048.2078186}
Collobert, R., Weston, J., Bottou, L., Karlen, M., Kavukcuoglu, K., Kuksa, P.:
\newblock Natural language processing (almost) from scratch.
\newblock J. Mach. Learn. Res. \textbf{12} (2011)

\bibitem{demori08-SPM}
De~Mori, R., Bechet, F., Hakkani-Tur, D., McTear, M., Riccardi, G., Tur, G.:
\newblock Spoken language understanding: A survey.
\newblock IEEE Signal Processing Magazine \textbf{25} (2008)  50--58

\bibitem{Sutskever-2014-SSL-2969033.2969173}
Sutskever, I., Vinyals, O., Le, Q.V.:
\newblock Sequence to sequence learning with neural networks.
\newblock In: Proceedings of NIPS, Cambridge, MA, USA, MIT Press (2014)

\bibitem{DBLP-journals-corr-BahdanauCB14}
Bahdanau, D., Cho, K., Bengio, Y.:
\newblock Neural machine translation by jointly learning to align and
  translate.
\newblock CoRR \textbf{abs/1409.0473} (2014)

\bibitem{Collins-1997-TGL}
Collins, M.:
\newblock Three generative, lexicalised models for statistical parsing.
\newblock In: Proceedings of ACL, Stroudsburg, PA, USA, Association for
  Computational Linguistics (1997)  16--23

\bibitem{Soon2001}
Soon, W.M., Ng, H.T., Lim, D.C.Y.:
\newblock A {M}achine {L}earning {A}pproach to {C}oreference {R}esolution of
  {N}oun {P}hrases.
\newblock Computational Linguistics \textbf{27} (2001)  521--544

\bibitem{Ng2002}
Ng, V., Cardie, C.:
\newblock Improving {M}achine {L}earning {A}pprocahes to {C}orefrence
  {R}esolution.
\newblock In: Proceedings of ACL'02. (2002)  104--111

\bibitem{Grouin.etAL:I2B2:2011}
Grouin, C., Dinarelli, M., Rosset, S., Wisniewski, G., Zweigenbaum, P.:
\newblock Coreference resolution in clinical reports. the limsi participation
  in the i2b2/va 2011 challenge.
\newblock In: In Proceedings of i2b2/VA 2011 Coreference Resolution Workshop.
  (2011)

\bibitem{dinarelli2012-eacl}
Dinarelli, M., Rosset, S.:
\newblock Tree representations in probabilistic models for extended named
  entity detection.
\newblock In: European Chapter of the Association for Computational Linguistics
  (EACL), Avignon, France (2012)  174--184

\bibitem{DINARELLI-ROSSET:OCR-NER:LREC2012}
Dinarelli, M., Rosset, S.:
\newblock Tree-structured named entity recognition on ocr data: Analysis,
  processing and results.
\newblock In Chair), N.C.C., Choukri, K., Declerck, T., Dogan, M.U., Maegaard,
  B., Mariani, J., Odijk, J., Piperidis, S., eds.: Proceedings of the Eight
  International Conference on Language Resources and Evaluation (LREC'12),
  Istanbul, Turkey, European Language Resources Association (ELRA) (2012)

\bibitem{Rush-2012-IPP-2390948.2391112}
Rush, A.M., Reichart, R., Collins, M., Globerson, A.:
\newblock Improved parsing and pos tagging using inter-sentence consistency
  constraints.
\newblock In: Proceedings of EMNLP-CoNLL, Stroudsburg, PA, USA (2012)

\bibitem{D17-1018}
Lee, K., He, L., Lewis, M., Zettlemoyer, L.:
\newblock End-to-end neural coreference resolution.
\newblock In: Proceedings of EMNLP, Association for Computational Linguistics
  (2017)

\bibitem{lample2016neural}
Lample, G., Ballesteros, M., Subramanian, S., Kawakami, K., Dyer, C.:
\newblock Neural architectures for named entity recognition.
\newblock arXiv preprint arXiv:1603.01360 (2016)

\bibitem{Ma-Hovy-ACL-2016}
Ma, X., Hovy, E.:
\newblock End-to-end sequence labeling via bi-directional lstm-cnns-crf.
\newblock In: Proceedings of ACL. (2016)

\bibitem{kemker2018measuring}
Kemker, R., McClure, M., Abitino, A., Hayes, T.L., Kanan, C.:
\newblock Measuring catastrophic forgetting in neural networks.
\newblock In: Thirty-Second AAAI Conference on Artificial Intelligence. (2018)

\bibitem{N18-1172}
Augenstein, I., Ruder, S., S{\o}gaard, A.:
\newblock Multi-task learning of pairwise sequence classification tasks over
  disparate label spaces.
\newblock In: Proceedings of the 2018 Conference of the North American Chapter
  of the Association for Computational Linguistics: Human Language
  Technologies, Volume 1 (Long Papers), Association for Computational
  Linguistics (2018)  1896--1906

\bibitem{DBLP-journals-corr-VinyalsKKPSH14}
Vinyals, O., Kaiser, L., Koo, T., Petrov, S., Sutskever, I., Hinton, G.E.:
\newblock Grammar as a foreign language.
\newblock CoRR \textbf{abs/1412.7449} (2014)

\bibitem{dinarelli_hal-01553830}
Dinarelli, M., Vukotic, V., Raymond, C.:
\newblock {Label-dependency coding in Simple Recurrent Networks for Spoken
  Language Understanding}.
\newblock In: {Interspeech}, Stockholm, Sweden (2017)

\bibitem{werbos-bptt}
Werbos, P.:
\newblock Backpropagation through time: what does it do and how to do it.
\newblock In: Proceedings of IEEE. Volume~78. (1990)  1550--1560

\bibitem{Hochreiter-1997-LSTM}
Hochreiter, S., Schmidhuber, J.:
\newblock Long short-term memory.
\newblock Neural Comput. \textbf{9} (1997)  1735--1780

\bibitem{Cho-2014-GatedRecurrentUnits}
Cho, K., van Merrienboer, B., G{\"{u}}l{\c{c}}ehre, {\c{C}}., Bougares, F.,
  Schwenk, H., Bengio, Y.:
\newblock Learning phrase representations using {RNN} encoder-decoder for
  statistical machine translation.
\newblock CoRR \textbf{abs/1406.1078} (2014)

\bibitem{46201}
Vaswani, A., Shazeer, N., Parmar, N., Uszkoreit, J., Jones, L., Gomez, A.N.,
  Kaiser, L., Polosukhin, I.:
\newblock Attention is all you need.
\newblock (2017)

\bibitem{Vukotic.etal_2016}
Vukotic, V., Raymond, C., Gravier, G.:
\newblock {A step beyond local observations with a dialog aware bidirectional
  GRU network for Spoken Language Understanding}.
\newblock In: {Interspeech}, San Francisco (2016)

\bibitem{LSTM-CNN-NER-2015}
Chiu, J.P.C., Nichols, E.:
\newblock Named entity recognition with bidirectional lstm-cnns.
\newblock CoRR \textbf{abs/1511.08308} (2015)

\bibitem{huang2015bidirectional}
Huang, Z., Xu, W., Yu, K.:
\newblock Bidirectional lstm-crf models for sequence tagging.
\newblock arXiv preprint arXiv:1508.01991 (2015)

\bibitem{Dupont-etAl-LDRNN-CICling2017}
Dupont, Y., Dinarelli, M., Tellier, I.:
\newblock Label-dependencies aware recurrent neural networks.
\newblock In: Proceedings of CICling, Budapest, Hungary, LNCS, Springer (2017)

\bibitem{2016:arXiv:DinarelliTellier:NewRNN}
Dinarelli, M., Tellier, I.:
\newblock Improving recurrent neural networks for sequence labelling.
\newblock CoRR \textbf{abs/1606.02555} (2016)

\bibitem{DinarelliTellier:RNN:CICling2016}
Dinarelli, M., Tellier, I.:
\newblock New recurrent neural network variants for sequence labeling.
\newblock In: Proceedings of the 17th International Conference on Intelligent
  Text Processing and Computational Linguistics, Konya, Turkey, Lecture Notes
  in Computer Science (Springer) (2016)

\bibitem{NIPS2017_6775}
Xia, Y., Tian, F., Wu, L., Lin, J., Qin, T., Yu, N., Liu, T.Y.:
\newblock Deliberation networks: Sequence generation beyond one-pass decoding.
\newblock In Guyon, I., Luxburg, U.V., Bengio, S., Wallach, H., Fergus, R.,
  Vishwanathan, S., Garnett, R., eds.: Advances in Neural Information
  Processing Systems 30.
\newblock Curran Associates, Inc. (2017)  1784--1794

\bibitem{Bonneau-Maynard2006-media}
Bonneau-Maynard, H., Ayache, C., Bechet, F., Denis, A., Kuhn, A., Lef\`evre,
  F., Mostefa, D., Qugnard, M., Rosset, S., Servan, S.~Vilaneau, J.:
\newblock Results of the french evalda-media evaluation campaign for literal
  understanding.
\newblock In: LREC, Genoa, Italy (2006)  2054--2059

\bibitem{Marcus93buildinga}
Marcus, M.P., Santorini, B., Marcinkiewicz, M.A.:
\newblock Building a large annotated corpus of english: The penn treebank.
\newblock COMPUTATIONAL LINGUISTICS \textbf{19} (1993)

\bibitem{DBLP-journals-corr-KimDHR17}
Kim, Y., Denton, C., Hoang, L., Rush, A.M.:
\newblock Structured attention networks.
\newblock CoRR \textbf{abs/1702.00887} (2017)

\bibitem{simonnet-hal-01433202}
Simonnet, E., Camelin, N., Del{\'e}glise, P., Est{\`e}ve, Y.:
\newblock {Exploring the use of Attention-Based Recurrent Neural Networks For
  Spoken Language Understanding}.
\newblock In: {SLUNIPS}, Montreal, Canada (2015)

\bibitem{Elman90findingstructure}
Elman, J.L.:
\newblock Finding structure in time.
\newblock COGNITIVE SCIENCE \textbf{14} (1990)

\bibitem{jordan-serial}
Jordan, M.I.:
\newblock Serial order: {A} parallel, distributed processing approach.
\newblock (1989)

\bibitem{Bengio-1994-RNN-Learning-Difficulty}
Bengio, Y., Simard, P., Frasconi, P.:
\newblock Learning long-term dependencies with gradient descent is difficult.
\newblock Trans. Neur. Netw. \textbf{5} (1994)  157--166

\bibitem{Mesnil-RNN-2015}
Mesnil, G., Dauphin, Y., Yao, K., Bengio, Y., Deng, L., Hakkani-Tur, D., He,
  X., Heck, L., Tur, G., Yu, D., Zweig, G.:
\newblock Using recurrent neural networks for slot filling in spoken language
  understanding.
\newblock IEEE/ACM TASLP (2015)

\bibitem{Bengio03aneural}
Bengio, Y., Ducharme, R., Vincent, P., Jauvin, C.:
\newblock A neural probabilistic language model.
\newblock JOURNAL OF MACHINE LEARNING RESEARCH \textbf{3} (2003)  1137--1155

\bibitem{raymond07-luna}
Raymond, C., Riccardi, G.:
\newblock Generative and discriminative algorithms for spoken language
  understanding.
\newblock In: Interspeech, Antwerp, Belgium (2007)

\bibitem{dinarelli09:Interspeech}
Dinarelli, M., Moschitti, A., Riccardi, G.:
\newblock Concept segmentation and labeling for conversational speech.
\newblock In: Proceedings of the International Conference of the Speech
  Communication Assosiation (Interspeech), Brighton, U.K. (2009)

\bibitem{Hahn.etAL-SLUJournal-2010}
Hahn, S., Dinarelli, M., Raymond, C., Lef\`evre, F., Lehen, P., De~Mori, R.,
  Moschitti, A., Ney, H., Riccardi, G.:
\newblock Comparing stochastic approaches to spoken language understanding in
  multiple languages.
\newblock IEEE TASLP \textbf{99} (2010)

\bibitem{Dinarelli2010.PhDThesis}
Dinarelli, M.:
\newblock Spoken Language Understanding: from Spoken Utterances to Semantic
  Structures.
\newblock PhD thesis, International Doctoral School in Information and
  Communication Technology, Dipartimento di Ingegneria e Scienza dell'
  Informazione, via Sommarive 14, 38100 Povo di Trento (TN), Italy (2010)

\bibitem{dinarelli2011:emnlp}
Dinarelli, M., Rosset, S.:
\newblock Hypotheses selection criteria in a reranking framework for spoken
  language understanding.
\newblock In: Conference of Empirical Methods for Natural Language Processing,
  Edinburgh, U.K. (2011)  1104--1115

\bibitem{Dinarelli.etAl-SLU-RR-2011}
Dinarelli, M., Moschitti, A., Riccardi, G.:
\newblock Discriminative reranking for spoken language understanding.
\newblock IEEE TASLP \textbf{20} (2011)  526--539

\bibitem{Vapnik98-book}
Vapnik, V.N.:
\newblock {S}tatistical {L}earning {T}heory.
\newblock John Wiley and Sons (1998)

\bibitem{lafferty01-crf}
Lafferty, J., McCallum, A., Pereira, F.:
\newblock Conditional random fields: Probabilistic models for segmenting and
  labeling sequence data.
\newblock In: Proceedings of ICML, Williamstown, USA (2001)

\bibitem{RNNforSLU-Interspeech-2013}
Mesnil, G., He, X., Deng, L., Bengio, Y.:
\newblock Investigation of recurrent-neural-network architectures and learning
  methods for spoken language understanding.
\newblock In: Interspeech. (2013)

\bibitem{RNNforLU-Interspeech-2013}
Yao, K., Zweig, G., Hwang, M.Y., Shi, Y., Yu, D.:
\newblock Recurrent neural networks for language understanding, Interspeech
  (2013)

\bibitem{Vukotic.etal_2015}
Vukotic, V., Raymond, C., Gravier, G.:
\newblock Is it time to switch to word embedding and recurrent neural networks
  for spoken language understanding?
\newblock In: InterSpeech. (2015)

\bibitem{SLUwithLSTM-NN-IEEEWshop-2014}
Yao, K., Peng, B., Zhang, Y., Yu, D., Zweig, G., Shi, Y.:
\newblock Spoken language understanding using long short-term memory neural
  networks, IEEE (2014)

\bibitem{2017-NoRNN}
Wang, C.:
\newblock Network of recurrent neural networks.
\newblock CoRR \textbf{abs/1710.03414} (2017)

\bibitem{Holland-1999-ECO-520475}
Holland, J.H.:
\newblock Emergence: From Chaos to Order.
\newblock Perseus Publishing (1999)

\bibitem{RePEc-wop-safiwp-93-11-070}
Arthur, W.B.:
\newblock On the evolution of complexity.
\newblock Working papers, Santa Fe Institute (1993)

\bibitem{P18-2116}
Levy, O., Lee, K., FitzGerald, N., Zettlemoyer, L.:
\newblock Long short-term memory as a dynamically computed element-wise
  weighted sum.
\newblock In: Proceedings of ACL. (2018)

\bibitem{Dehghani2018UniversalT}
Dehghani, M., Gouws, S., Vinyals, O., Uszkoreit, J., Kaiser, L.:
\newblock Universal transformers.
\newblock CoRR \textbf{abs/1807.03819} (2018)

\bibitem{2018-ijcai-lstm-ldrnn}
Zhang, Y., Chen, H., Zhao, Y., Liu, Q., Yin, D.:
\newblock Learning tag dependencies for sequence tagging.
\newblock In: International Joint Conference on Artificial Intelligence
  (IJCAI). (2018)

\bibitem{Ramshaw95-BIO}
Ramshaw, L., Marcus, M.:
\newblock Text chunking using transformation-based learning.
\newblock In: Proceedings of the 3rd Workshop on Very Large Corpora, Cambridge,
  MA, USA (1995)  84--94

\bibitem{LeakyReLU-PReLU-2015}
He, K., Zhang, X., Ren, S., Sun, J.:
\newblock Delving deep into rectifiers: Surpassing human-level performance on
  imagenet classification.
\newblock In: 2015 {IEEE} International Conference on Computer Vision, {ICCV}
  2015, Santiago, Chile, December 7-13, 2015. (2015)  1026--1034

\bibitem{PracticalRecommendations-Bengio-2012}
Bengio, Y.:
\newblock Practical recommendations for gradient-based training of deep
  architectures.
\newblock CoRR \textbf{abs/1206.5533} (2012)

\bibitem{paszke2017automatic}
Paszke, A., Gross, S., Chintala, S., Chanan, G., Yang, E., DeVito, Z., Lin, Z.,
  Desmaison, A., Antiga, L., Lerer, A.:
\newblock Automatic differentiation in pytorch.
\newblock In: NIPS-W. (2017)

\bibitem{NIPS2016_6221}
Gruslys, A., Munos, R., Danihelka, I., Lanctot, M., Graves, A.:
\newblock Memory-efficient backpropagation through time.
\newblock (2016)  4125--4133

\bibitem{Yeh00moreaccurate}
Yeh, A.:
\newblock More accurate tests for the statistical significance of result
  differences.
\newblock  (In: Proceedings of Coling)

\bibitem{sigf06}
Pad\'o, S.:
\newblock User's guide to \texttt{sigf}: Significance testing by approximate
  randomisation. (2006)

\bibitem{pennington2014glove}
Pennington, J., Socher, R., Manning, C.D.:
\newblock Glove: Global vectors for word representation.
\newblock In: Empirical Methods in Natural Language Processing (EMNLP). (2014)
  1532--1543

\end{thebibliography}
\end{document}